# FORECAST-CLSTM: A New Convolutional LSTM Network for Cloudage Nowcasting

Chao Tan[*], Xin Feng[*], Jianwu Long[*], Li Geng[#]

[*]College of Computer Science and Engeering, Chongqing University of Technology,Chongqing ,China
[#]New York City College of Technology of City University of New York, Brooklyn, NY, USA
Email: istvartan@outlook.com, {xfeng,jwlong}@cqut.edu.cn, Lgeng@citytech.cuny.edu

*Abstract*— **With the highly demand of large-scale and real-time weather service for public, a refinement of short-time cloudage prediction has become an essential part of the weather forecast productions. To provide a weather-service-compliant cloudage nowcasting, in this paper, we propose a novel hierarchical Convolutional Long-Short-Term Memory network based deep learning model, which we term as FORECAST-CLSTM, with a new Forecaster loss function to predict the future satellite cloud images. The model is designed to fuse multi-scale features in the hierarchical network structure to predict the pixel value and the morphological movement of the cloudage simultaneously. We also collect about 40K infrared satellite nephograms and create a large-scale Satellite Cloudage Map Dataset(SCMD). The proposed FORECAST-CLSTM model is shown to achieve better prediction performance compared with the state-of-the-art ConvLSTM model and the proposed Forecaster Loss Function is also demonstrated to retain the uncertainty of the real atmosphere condition better than conventional loss function.**

## I. Introduction

In weather forecasting, a mass of quantitative data about the current state of the atmosphere at a given place is collected and meteorologists use meteorology methods to analyse and predict the change of the atmosphere. Cloudage information is an important meteorological factor that can reflects the current rainfall and air distribution. Effective cloudage nowcasting would help to generate the ultraviolet airing index, measure the Land Surface Temperature and provide weather guidance for astronomical observation. Hence, to make an accurate short time cloudage prediction is a significant task in the field of short range weather forecasting. In current weather forecasting, weatherman combines the result of the Numerical Weather Analyse(NWA) of some related meteorological elements, e.g. air pressure, wind direction, etc., with self-experiences to measure the short time cloudage variation in a local area [1–4]. However, these meteorological methods can not make accurate and timely prediction for large area cloudage nowcasting. Therefore, the demands of real-time, fine-grained and large-scale weather prediction leverage cloudage nowcasting to be a challenging task in the meteorological community.

Long-Short-Term Memory[5](LSTM) network is one of the most successful models in learning the long-term dependencies of dynamic sequences for recent computer vision works such as natural language processing[6], video frames prediction[7], pathologic prediction[8], etc. However, this conventional LSTM network does not take spatial information into account for image sequence prediction task. Recently, [9] proposed a new convolutional LSTM(CLSTM) network, where the fully connections in state-to-state and input-to-state transitions of the conventional LSTM layer are replaced by the convolution operations. Many successive studies have shown that CLSTM could achieve better performance in handling the spatio-temporal correlations for spatial sequence data prediction. For example, Shi [9] and Kim [10] use CLSTM to solve the precipitation nowcasting problem, Bates [11] takes a CLSTM based model to extract 3D vascular structures from microscopy images and Luo [12] uses a CLSTM structure to encode the variations of appearance and motion for normal events in the task of abnormal video event detection. These works show that CLSTM is an ideal model for solving the spatio-temporal prediction tasks of multi-dimension sequence data.

In this paper, we present a novel hierarchical Convolution Long-Short-Term Memory based model trained with a proposed forecaster loss，which is termed as FORECAST-CLSTM, to further enable the spatial coding ability of CLSTM and make more reasonable cloudage prediction. On the other hand, we also create the SCMD16 dataset that contains 38,000 sequences of satellite nephogram sequences. To the best of our knowledge, this is the first large scale, fine-grain cloudage nowcasting data collection and we are the first to use the actual weather forecaster metric to construct an end-to-end CLSTM model to fulfill the real-time, large-scale cloudage nowcasting task.

## II. FORECAST-CLSTM Model

### A. Conventional CLSTM Model

A typical CLSTM model is a simple auto-encoder structure which is stacked by multiple CLSTM and pooling layers. Within each CLSTM, the task of spatial feature extraction and temporal prediction is performed simultaneously. Compared to the conventional LSTM model, the number of hidden units for CLSTM layer represents the channels of the hidden feature maps and every connection in CLSTM layer is connected by convolution operation.

*B. FORECAST-CLSTM Model*

The CLSTM model is capable of modeling the spatio-temporal sequence data. However, because of the stack of multiple CLSTM layers, the computation complexity of the network is extremely large. By looking into the network structure, there are two different types of convolution kernels for each gate unit in CLSTM layers. One convolution kernel is used to make convolution from input-to-state, the other is to deal with the convolution of state-to-state transition. Although the purpose of the CLSTM structure is to leverage the network to have the spatial feature perception and temporal prediction ability simultaneously, there are considerable redundant convolution operations.

Motivated by this observation, to better extract high level spatial features and reduce the redundant computation of the CLSTM model, a natural idea is to separate the spatial feature extraction work from the CLSTM model, and use the better spatial feature extraction structure, i.e. Convolutional Neural Network(CNN) instead. We can thus construct a network by stacking multiple efficient frame by frame convolution and pooling layers alternately at the front-end of the model and then pass the advanced features directly to a single CLSTM layer to get the final prediction result. This structure would hence lead to a significant decline in parameter and computation complexity compared with conventional CLSTM model. However, this kind of network makes the temporal variation at the last step of the model, which implies that the temporal information of low-level features are ignored. That is to say, the low-level features in the temporal prediction are considered as relatively constant. This hypothesis is reasonable for most natural scene prediction task, but maybe problematic in handling the nephogram data in the following two aspects: the first problem is that the semantic concept of cloudage feature is uncertain. Compared with natural scene images, such as streetscape and portrait images, the pixel value in the nephogram is more like random distributed, and do not have some certain motion patterns. It's hence difficult to abstract the semantic feature from low-level network layers to the high-level layers. Secondly, in the real atmosphere, the spatial location transform and the morphological changes of the cloud occur simultaneously. However, when we learn to predict the cloudage moving in the deep neural network model, the higher level abstract feature usually encode the overall position, and most inner morphological changes of the clouds are captured by the small-scale low-level features. Therefore, we need to include both the high-level and low-level features to make more reasonable and accurate prediction for the unnatural scene prediction problems.

Here, we propose a hierarchical CLSTM based model called FORECAST-CLSTM (which will be denoted as F-CLSTM model hereafter) to add several layers of sub-encoder networks to encode different scales of feature maps, the error of the whole model will thus be distributed over multi-scale feature maps.

Specifically, we use a sequence convolution (seq-CONV) layer followed by a convolution LSTM (CLSTM) layer for image feature extraction and sequence prediction respectively (SeqConv-CLSTM). Except for the last SeqConv-CLSTM structure, a maxpooling operation is all applied to reduce the dimension of the feature map that is produced by convolution LSTM layer (Conv-CLSTM-POOLING). Meanwhile, for the prediction of each convolution LSTM layer, a predicted image of this feature scale is carried out by using a decoder network which is completely symmetric with the current pervious encoding network consisting of deconvolution and upper sampling layers. A typical F-CLSTM model is a stack of SeqConv-CLSTM and CONV-CLSTM-POOLING structures. The loss of the whole network is the sum of the error of different scales of prediction and the output of the F-CLSTM model is the fusion of the prediction results of different feature levels. The framework of the F-CLSTM model is illustrated in Figure 1.

Generally, the F-CLSTM model takes the spatial and temporal variations of multi-scale features into account. And the fusion of multi-losses from each sub-decoder will not suffer from much information loss caused by multiple pooling layers.

### III. FORECASTER METRIC

In the real atmosphere, the atmospheric movement is a highly chaotic system. Due to this high uncertainty, it is impossible for us to predict with a 100% accuracy. In current actual weather forecast work, the weatherman generally combines numerical computing results with the reality conditions to derive a relative 'safe' prediction. Hence, to be accordance with the weather-service-compliant prediction, the cloudage nowcasting system should provide an accurate predict under the premise of retaining as much uncertainty of such fuzzy pixels as possible.

An obvious manifestation of the uncertainty in our models is the blurring of the prediction image. The reason for blurring is not only the inherent chaos of the atmosphere but also the calculation of the loss function. One significant trait of the cloudage nowcasting task is that the uncertainty of the prediction is more likely to occur in the low cloudage areas, such as the edge of the clouds and the beginning or dissipation of convection. However, by using the mean square error, the error for every pixel between prediction and ground truth is equally considered, thus the uncertainty is randomly dispersed and blurring is caused throughout the image. Therefore, in order to generate a weather-service-compliant prediction, the low cloudage areas are assumed to contribute more errors due to their nature of uncertainty.

We propose a mean square error based loss function for cloudage nowcasting named Forecaster Loss Function. For each output image sample at time t, the Forecaster loss between the predicted nephogram $\hat{y}_{ij}^t$ and the ground truth $y_{ij}^t$ is:

$$loss^t = \sum_{i=1}^{M} \sum_{j=1}^{N} e^{1-\frac{\min(|y_{ij}^t|,|\hat{y}_{ij}^t|)}{255}} \times (y_{ij}^t - \hat{y}_{ij}^t)^2 \quad (1)$$

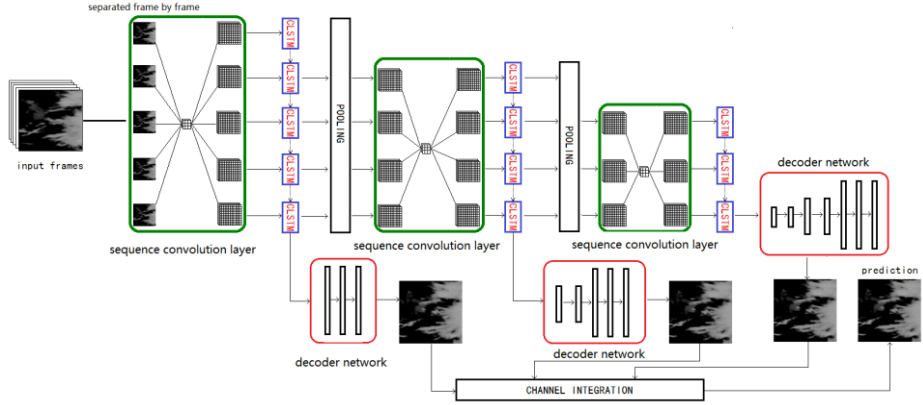

Fig. 1 Architecture of the FORECAST-CLSTM Model. The green box represents the sequence convolution layer and the red box represents the decoding network that is completely symmetric with the previous structure consists of deconvolution and upper sampling layer. The channel integration structure is composed of a stack of multiple convolution layers with a kernel size of $1\times 1$.

and $|y_{ij}^t|, |\hat{y}_{ij}^t|$ denotes the absolute value of $y_{ij}^t$ and $\hat{y}_{ij}^t$. Unlike MSE, forecaster loss assigns higher weights to pixels with lower value by using a nonlinear function with multiple punishment rates.

## IV. EXPERIMENTS AND RESULT ANALYSIS

### A. SCMD16 Dataset

In the paper, we collect the infrared satellite nephograms of Asia region obtained by the FY-2F meteorological satellite of China via the Internet and create a large-scale Satellite Cloudage Map Dataset(SCMD16). To get rid of the impact of seasonal differences, we collect a total of 21811 nephograms from January 1, 2016 to December 31st. Each nephogram is a 3-channels RGB image of range from 0 to 255 with a size of 1100×1700. We labelled each nephogram with the corresponding time and make sequences for each of the six adjacent nephograms (30 minutes time interval between each two nephograms). In the collected nephograms, because on much of the days and regions of the images are just background or very close to background, we randomly cropped the nephograms into sub-images with spatial dimension of 200×200, and finally create a dataset that contains 35,000 nephogram sequences as training data and 3,000 sequences for testing.

Before training, for every frame in the SCMD16 dataset, background subtraction is first applied. We then convert the RGB images into grayscale and each sequence is randomly shuffled. For every experiment, we use 6 frames of nephograms to predict the next frame. All the models in the experiment are implemented using the open source PyTorch library. The weights in the models were initialized according to uniform initialization and the training was done using Nesterov Adam with a learning rate of 0.002. We run all the experiments on a computer with one NVIDIA GTX1080 Ti GPU.

### B. Experiment on Moving Mnist++ Dataset

Moving Mnist++ is a synthetic dataset and has been widely used as the standard dataset for image sequence prediction task. Here, we follow the generation process in [13] to generate the Moving Mnist sequences. All the sequence samples in Moving Mnist++ dataset are 10 frames long (we use 9 frames as inputs and one frame for prediction), each sequence contains two handwritten digits which are chosen randomly in Mnist dataset bouncing inside a 64×64 patch. In addition, digits in the patch are allowed to have random rotations, scale changes, and illumination changes respectively.

We evaluate the performance of F-CLSTM and conventional CLSTM model on Moving Mnist++ dataset. In the F-CLSTM model, we have designed two SeqConv-CLSTM-MAXPOOLING structure and single SeqConv--CLSTM architecture to encode and predict the image sequences. The channel number of each SeqConv and CLSTM layer is set to 32,32,64,64,128, and 128, and the kernel size used in the layer above is 3×3. In addition, we use a hidden 1×1 convolution layer containing 16 feature channels in the channel integration operation to generate the final prediction result. For CLSTM model, we use three convolutional LSTM layer with 32, 64 and 128 hidden channels respectively and two maxpooling layers.

In this paper, each convolution operation is followed by the Batch Normalization layer and all the activation functions are RELU. The result of the experiments and some of prediction sample images are shown in Table I and Figure 2.

### C. Experiment on SCMD2016 Dataset

We further evaluate the performance of the proposed F-CLSTM、CLSTM model and some baseline models including CLSTM, FC-LSTM[21] and multi-layer perceptron(MLP) on the SCMD2016 Dataset. The implementation of these methods are as following:

1) F-CLSTM: the same architechture as it on Moving Mnist++ dataset. The channel number of each Seq-CONV and CLSTM layer is set to 16,16,32,32,64 and 64. The kernel size of each layer is 3×3.
2) MLP: a two-layer neural network with 256 cells for each hidden layer.
3) FC-LSTM : a single LSTM layer with 256 hidden units
4) CLSTM: three convolutional LSTM layer with 16, 32 and 64 hidden channels respectively, and two maxpooling layers.

The result of the performance evaluation and some of prediction sample images are shown in Table I and Figure 3.

TABLE I
PERFORMANCE OF DIFFERENT MODELS

| Experiments on Moving Mnist++ Dataset | | | |
|---|---|---|---|
| **Model** | **MSE** | **PSNR** | **SSIM** |
| CLSTM | 221.7520 | 22.3217 | 0.8106 |
| F-CLSTM | **70.9074** | **31.6401** | **0.9620** |
| Experiments on SCMD16 Dataset | | | |
| **Model** | **MSE** | **PSNR** | **SSIM** | **ECCR** |
| MLP | 377.2251 | 24.0012 | 0.9305 | - |
| FC-LSTM | 359.4351 | 24.5127 | 0.9377 | - |
| CLSTM | 143.1854 | 27.6126 | 0.9744 | - |
| F-CLSTM | **113.2776** | **29.6724** | **0.9800** | 54.8% |
| F-CLSTM-F | 114.2779 | 29.6578 | 0.9798 | **55.9%** |

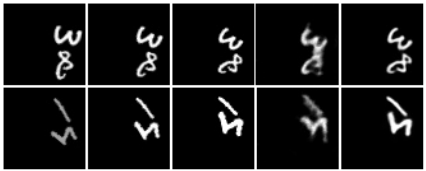

Fig 2. Two predicted moving mnist++ images using different prediction models. From left to right are the last two frames of the input sequences, ground truth images, prediction by CLSTM and F-CLSTM model respectively.

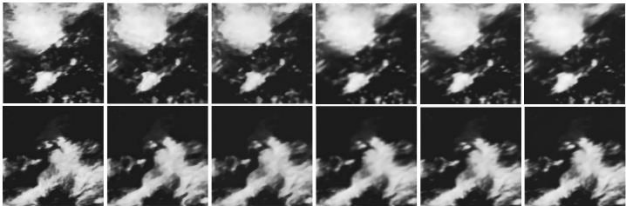

Fig 3. Two predicted cloudage images using different prediction models. From left to right are ground truth nephograms, prediction by FC-LSTM, MLP, CLSTM, F-CLSTM and F-CLSTM-F with Forecaster-loss respectively.

### D. Result Analysis

We first perform experiment on Moving Mnist++ dataset. We measure the MSE, Peak Signal to Noise Ratio (PSNR) and Structural Similarity Index(SSIM) criterions for every model on the testing data. According to the experimental result in Table I, because F-CLSTM considered the spatial-temporal variations of different feature levels, the performance is better than CLSTM model for all evaluation criterions. It can also be seen from Fig.2 that the prediction result of F-CLSTM is sharper and more accurate than that of CLSTM model.

On the other hand, because the conventional MLP and FC-LSTM model don not consider the spatial correlation of the nephogram, the performance of both models are not good in terms of pixel correlation. MLP assumes that every pixel in the nephogram only has global motion, and the cloud shape would keep unchanged during the short-time-period. While FC-LSTM considers individual movement for each pixel, but 6 frames is not sufficient for it to learn the good temporal representation. Comparatively, models that are based on convolutional LSTM lead to much better performance, which indicates that the combination of convolution and LSTM is crucial for spatial-temporal nephogram sequences. Compared with the conventional CLSTM models, the F-CLSTM model achieves higher prediction performance. It can be seen from Figure 3 that the F-CLSTM model could preserve more detailed features than CLSTM to obtain more accurate prediction result.

Finally, in order to measure the performance of MSE and the proposed Forecaster Loss, we designed an evaluation metric called Effective Cloud Coverage Ratio (ECCR). ECCR represents the ratio of the pixel with cloud presence in the image to the entire space. It can be considered that the larger ECCR, the 'safer' prediction made. From the last two rows in Table I, it can be seen that model with Forecaster loss can make a 'safer' prediction than traditional MSE loss function and does not sacrifice the prediction accuracy at the same time. In Table I, models ending in '-F' indicate that Forecaster loss is used during training.

## V. CONCLUSION

In this paper, we proposed a deep learning model called FORECAST-CLSTM for cloudage nowcasting task. We also designed a new weather-service- compliant forecaster loss function to make the prediction accordant with the actual weather casting. The performance of our models was investigated by extensive experiments on the moving mnist++ and collected infrared satellite nephogram dataset.